\documentclass[10pt,twocolumn,letterpaper]{article}

\usepackage{wacv}              

\usepackage{graphicx}
\usepackage{amsmath}
\usepackage{amssymb}
\usepackage{booktabs}
\usepackage{multirow}
\usepackage[accsupp]{axessibility}
\usepackage{xcolor} 

\usepackage[pagebackref,breaklinks,colorlinks]{hyperref}

\usepackage[capitalize]{cleveref}
\crefname{section}{Sec.}{Secs.}
\Crefname{section}{Section}{Sections}
\Crefname{table}{Table}{Tables}
\crefname{table}{Tab.}{Tabs.}

\def\wacvPaperID{760} 
\def\confName{WACV}
\def\confYear{2025}

\begin{document}

\title{Dance Any Beat: Blending Beats with Visuals in Dance Video Generation}

\author{Xuanchen Wang$^{1}$ \enskip Heng Wang$^{1}$ \enskip Dongnan Liu$^{1}$ \enskip Weidong Cai$^{1}$ \\
$^{1}$University of Sydney \\
{\tt\small xwan0579@uni.sydney.edu.au}\\
{\tt\small \{heng.wang, dongnan.liu, tom.cai\}@sydney.edu.au}
}


\twocolumn[{
\maketitle
\begin{center}
    \captionsetup{type=figure}
    \includegraphics[width=\textwidth]{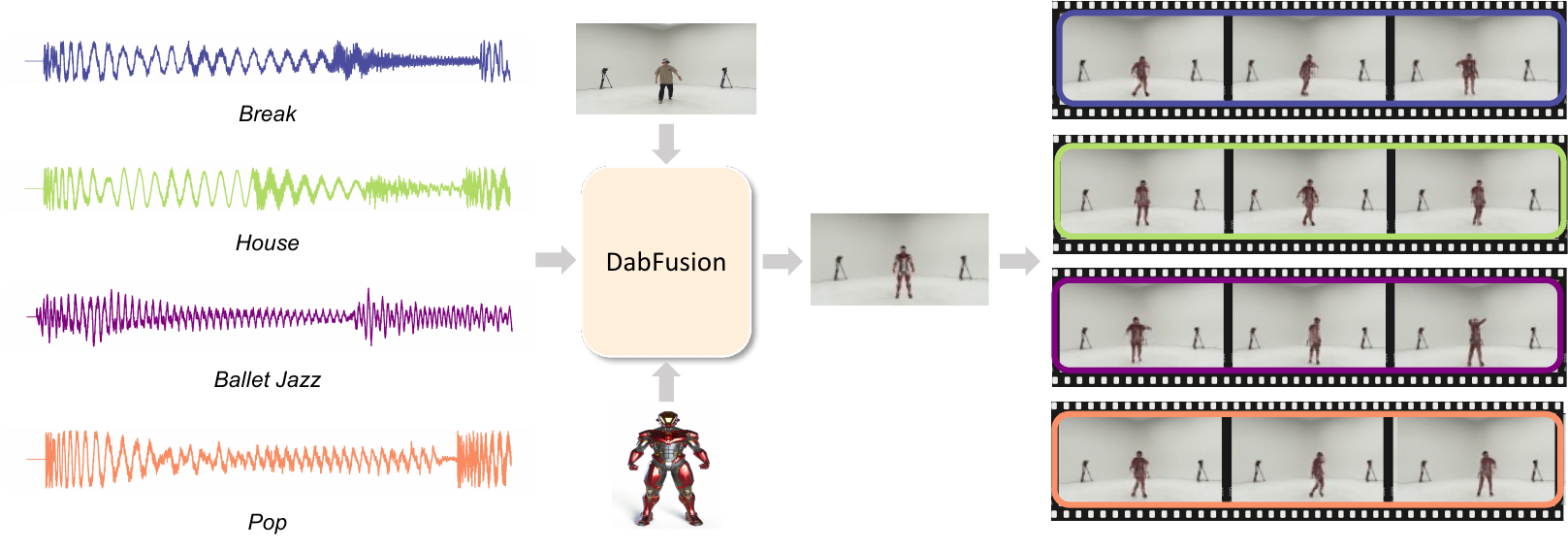}
    \captionof{figure}{We introduce DabFusion, a diffusion-based framework designed to generate videos of individuals dancing, utilizing a music input and an initial reference image as the conditions for video generation.}
    \label{fig:teaser}
\end{center}
}]

\begin{abstract}
Generating dance from music is crucial for advancing automated choreography. Current methods typically produce skeleton keypoint sequences instead of dance videos and lack the capability to make specific individuals dance, which reduces their real-world applicability. These methods also require precise keypoint annotations, complicating data collection and limiting the use of self-collected video datasets. To overcome these challenges, we introduce a novel task: generating dance videos directly from images of individuals guided by music. This task enables the dance generation of specific individuals without requiring keypoint annotations, making it more versatile and applicable to various situations. Our solution, the Dance Any Beat Diffusion model (DabFusion), utilizes a reference image and a music piece to generate dance videos featuring various dance types and choreographies. The music is analyzed by our specially designed music encoder, which identifies essential features including dance style, movement, and rhythm. DabFusion excels in generating dance videos not only for individuals in the training dataset but also for any previously unseen person. This versatility stems from its approach of generating latent optical flow, which contains all necessary motion information to animate any person in the image. We evaluate DabFusion's performance using the AIST++ dataset, focusing on video quality, audio-video synchronization, and motion-music alignment. We propose a 2D Motion-Music Alignment Score (2D-MM Align), which builds on the Beat Alignment Score to more effectively evaluate motion-music alignment for this new task. Experiments show that our DabFusion establishes a solid baseline for this innovative task. Video results can be found on our project page: \href{https://DabFusion.github.io}{https://DabFusion.github.io}.

\end{abstract}

\section{Introduction}
Music-to-dance generation not only revolutionizes choreography by automating complex routines but also enhances performance training, enriches interactive gaming experiences, and opens new avenues for virtual reality entertainment, offering profound benefits across multiple real-world applications. Existing methods \cite{li2020learning, li2021ai, siyao2022bailando, qi2023diffdance} focus on generating keypoint sequences, which, despite being effective, often produce outputs that are less intuitive and do not typically support the animation of specific individuals, which limits their practical utility in personalized applications. Ren et al. \cite{ren2020self} attempt to advance the field, but their approach still begins by generating keypoint sequences, which are then used to synthesize videos of specific individuals. This method requires an additional motion transfer model and a video of the target person, complicating the pipeline and limiting practical applications, as obtaining a video of the target person can be challenging in real-world scenarios. Another challenge with existing techniques is their dependency on precisely annotated keypoint data. This requirement not only complicates the data collection process but also restricts the usability of these methods with self-collected or less structured video datasets. To address these limitations, we introduce a new and innovative task: generating dance videos directly from images of individuals, guided by accompanying music. By utilizing images of individuals, this new task facilitates the generation of personalized dance sequences that can feature specific people. Furthermore, it eliminates the need of keypoint annotations, thereby increasing the versatility and applicability of the task across a variety of scenarios. Our solution, named the Dance Any Beat Diffusion model (DabFusion), takes a reference image paired with a selected piece of music as input to create dance videos featuring various dance types and choreographies. For extracting information from music, we develop a novel music encoder that captures essential features including dance style, movement, and rhythm, which are critical for this task. To accurately extract dance styles, we fine-tune the CLAP \cite{wu2023large} model, a text-audio pretraining foundation model, on the AIST++ \cite{li2021ai} dataset, which serves as the training dataset for our task. For movement information, we fine-tune the Wav2CLIP \cite{wu2022wav2clip} model, a visual-audio pretraining foundation model, on the AIST++ dataset. The synchronization of dance movements with the music's rhythm is essential for realistic dance generation. To achieve this, we employ Librosa \cite{mcfee2015librosa}, a tool for audio signal analysis, to extract beat information from the music. By integrating these elements, our music encoder can generate a comprehensive representation of the music. 

\begin{figure*}[t]
  \centering
  \includegraphics[width=\textwidth]{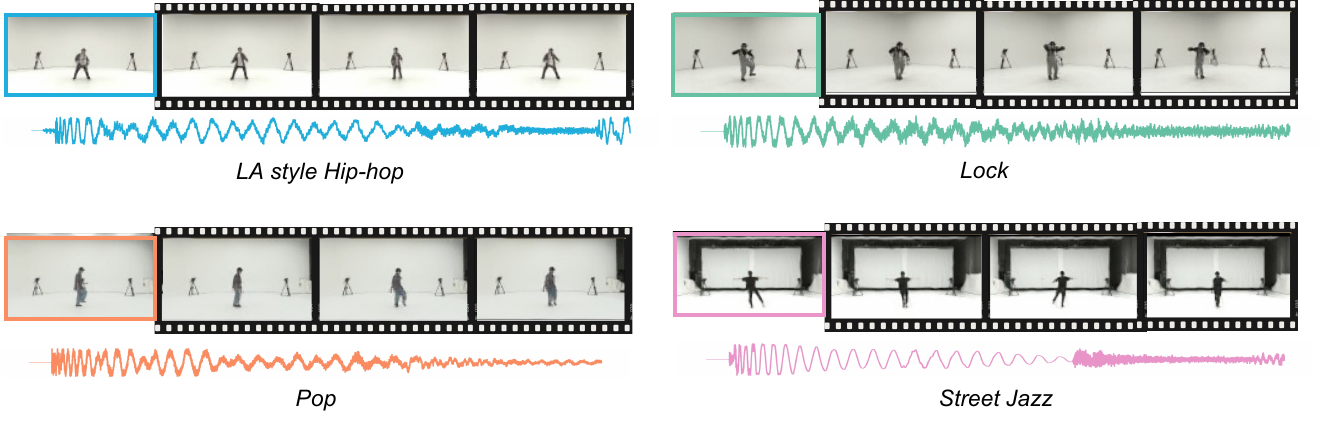}
  \caption{Exemplar videos generated from our DabFusion. Taking first image as starting frame and the unique music clip as guiding dance style, our framework is capable of generating varied styles of dance videos featuring different dancers from multiple perspectives with diverse initial poses and positions.}
  \label{fig:showcase}
\end{figure*}

DabFusion effectively generates dance videos not just for individuals included in the AIST++ dataset, but also for people it has never encountered before. As shown in \cref{fig:teaser}, we animate previously unseen individuals by merging their images with a background from the AIST++ dataset, bringing them to life through dance. DabFusion can choreograph dance sequences for previously unseen individuals to various music pieces, producing distinctly different videos. This diversity in output highlights the robust information extraction capability of our music encoder. When generating dance videos using the AIST++ dataset, as illustrated in \cref{fig:showcase}, DabFusion excels in producing varied dance styles with different dancers from multiple perspectives and diverse initial poses. The capability of DabFusion to generate various dance forms enables it to serve a broad audience, ranging from professional choreographers seeking inspiration to casual users wanting to see themselves or others dance to their favorite tunes. The versatility of DabFusion arises from its method of generating latent optical flow, which captures all essential motion information needed to animate any person in the image. This technique is widely adopted in the field of image-to-video (I2V) generation \cite{wang2020imaginator, rombach2022high, han2022show, Chan_2019_ICCV, wang2022latent, ren2020deep, siarohin2021motion, zhao2022thin}. Inspired by recent I2V works \cite{ni2023conditional, siarohin2019first, wiles2018x2face}, our model's training process is structured into two phases. First, a latent flow auto-encoder is trained unsupervisedly to estimate the latent optical flow between reference and driving frames in a video, aiding in warping the reference frame to generate movement. Then, a U-Net based diffusion model generates latent flows guided by music and a starting image. We conduct a thorough evaluation of DabFusion's performance using the AIST++ dataset, with particular focus on video quality, audio-video synchronization, and the alignment between motion and music. To more effectively assess motion-music alignment—a crucial aspect of this new task—we introduce a 2D motion-music alignment (2D-MM Align) score, inspired by the Beat Alignment Score \cite{li2021ai} for 3D motion-music correlation. This metric evaluates synchronization between motion and music in 2D scenarios, demonstrating the effectiveness of our approach in producing rhythmically aligned dance videos. Our experimental results demonstrate that DabFusion establishes a robust baseline for this innovative task, indicating its potential as a pioneering tool in the field of dance video synthesis.

Our contributions can be summarized as follows: 
\begin{itemize}
\item {}We introduce a novel task that directly generates dance videos from music, significantly enhancing the intuitiveness of the generated content and its applicability in real-world settings.
\item{} We propose DabFusion, a solution for the new task. DabFusion is capable of creating dance sequences not only for individuals present in the training dataset but also for person it has never previously encountered.
\item{} We propose a novel music encoder designed to generate a comprehensive representation of music, capturing key elements including dance style, movement, and rhythm.
\item{}We establish a solid baseline for the new task, supported by extensive evaluations on the AIST++ dataset.
\end{itemize}

\section{Related Works}
Generating dance sequences from music uniquely intersects motion synthesis \cite{aksan2019structured, butepage2017deep, hernandez2019human, holden2016deep} and music interpretation \cite{wu2022wav2clip, mcfee2015librosa, wu2023large}, aiming to create choreographed movements synchronized with input music. This extends beyond traditional motion synthesis, as choreographed movements are complex to animate. Early research \cite{shlizerman2018audio, zhuang2022music2dance, ferreira2021learning} focused on producing 2D dance sequences due to the availability of online dance videos and advances in 2D human pose estimation \cite{cao2017realtime}. However, 2D predictions lack expressiveness and applicability, prompting a shift to 3D dance generation. Recent methods leverage LSTMs \cite{tang2018dance, kao2020temporally, yalta2019weakly}, GANs \cite{lee2019dancing, sun2020deepdance, ginosar2019learning, kim2022brand}, and Transformers \cite{huang2020dance, li2021ai, siyao2022bailando, qi2023diffdance, ye2020choreonet} for 3D motion generation. The AIST++ dataset \cite{li2021ai}, a rich compilation of 3D motion data, has significantly advanced the field. Notable models include FACT \cite{li2021ai}, a full-attention-based cross-modal transformer using sequence-to-sequence learning for lifelike 3D dance sequences, and Bailando \cite{siyao2022bailando}, which combines a pose VQ-VAE with a Motion GPT for temporal coherence via actor-critic learning. While existing research focuses on generating motion sequences from music, Ren et al. \cite{ren2020self} synthesized videos using these sequences. Our work bypasses intermediary joint sequence generation, directly creating dance videos from music.

\begin{figure*}[t]
  \centering
  \includegraphics[scale = 0.6]{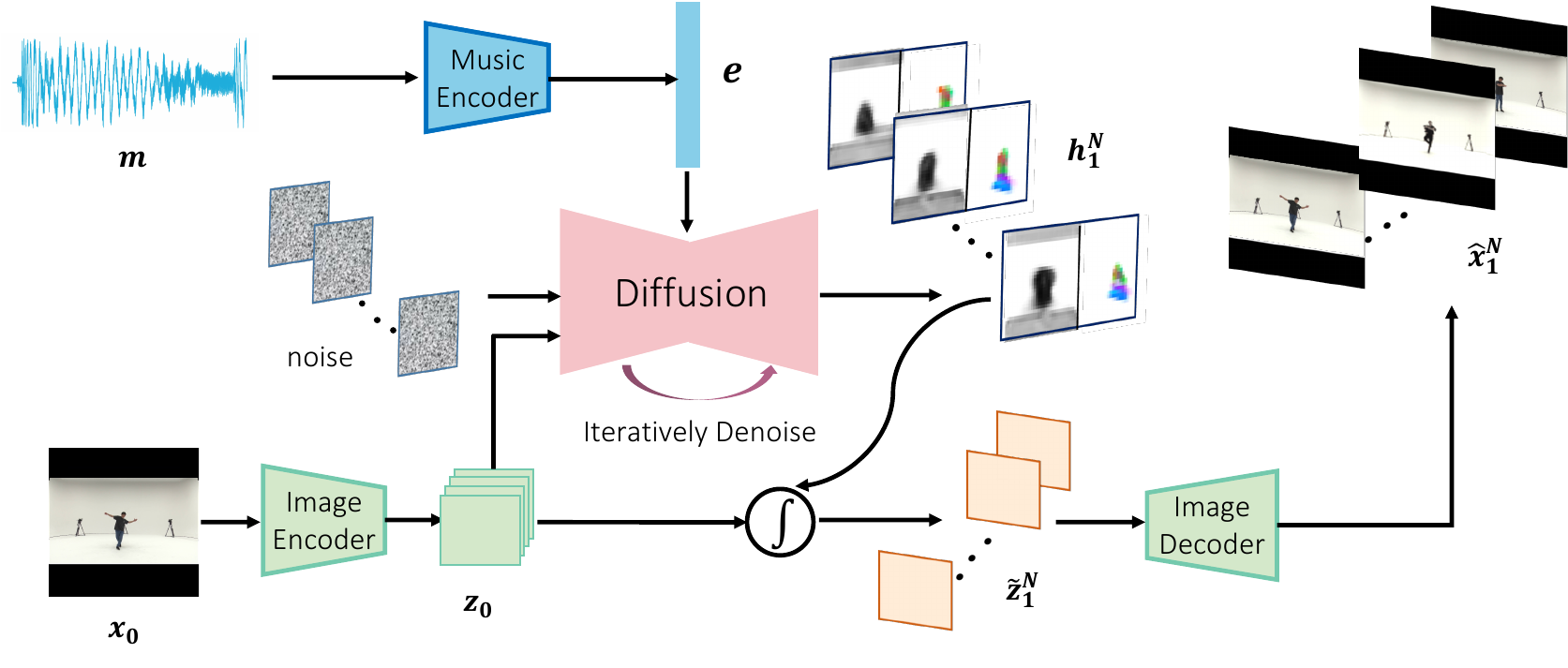}
  \caption{Overview of DabFusion. Given a reference image $ x_{0} $ with dimensions $ H_{x} \times W_{x} \times 3 $ and a piece of music $ m $. DabFusion incorporates noise input along with image embedding $ z_{0} $ which has dimensions $ H_{z} \times W_{z} \times C_{z} $ and music embedding $ e $ as conditions. Following the denoising stage of the diffusion model, we obtain $ h_{1}^{N} $ which has dimensions $ H_{z} \times W_{z} \times 3 \times N  $ , comprising a concatenated sequence of latent flow and corresponding occlusion maps. $ h_{1}^{N} $ is utilized to transform $ z_{0} $ into a new sequence of latent maps, denoted as $ \tilde{z}_{1}^{N} $, which is subsequently decoded to produce an image sequence.}
  \label{fig:stagetwo}
\end{figure*}

\section{Method}
\subsection{Overview}
Our methodology generates latent optical flows driven by musical inputs, with each music piece encoded by our specially designed music encoder. Music encoding is detailed in \cref{sec: music-encoding}. Recent works \cite{ni2023cross, siarohin2019first, wang2019few, wang2018video, ni2023conditional} in motion transfer have shown the effectiveness of using latent optical flow for warping images. This approach is more resource-efficient, requiring less computational power and time compared to high-dimensional pixel or latent feature spaces. Our methodology begins by training an auto-encoder to discern optical flow between video frames, we discuss this in details in \cref{sec: latent-estimation}. This trained auto-encoder then aids in training the diffusion model to generate latent flows. The training of the diffusion model is detailed in \cref{sec: diffusion-training}. The overview of DabFusion is shown in \cref{fig:stagetwo} , given an initial image $ x_{0} $ and a piece of music $m$, the image is first encoded into a latent space representation $ z_{0} $, while the music is transformed into an embedding $e$. Subsequently, a volume of randomly sampled Gaussian noise, with dimensions $ N \times H_{z} \times W_{z} \times 3 $, undergoes a gradual denoising process by the U-Net. This process yields $ h_{1}^{N} $ which is the concatenation of a latent flow sequence $ f_{1}^{N} $ and corresponding occlusion map sequence $ m_{1}^{N} $. $ z_{0} $ is then warped by each latent flow in the latent flow sequence and corresponding occlusion map in the occlusion map sequence to generate a new latent map sequence $ \tilde{z}_{1}^{N} $. These maps are sequentially fed into the image decoder, which synthesizes the frames of the new video.

\subsection{Music Encoding}
\label{sec: music-encoding}
Extracting meaningful information from music is essential for generating dance videos, where dance type, dancer movement, and rhythm play critical roles. For dance type information, we utilize the CLAP model, a pipeline for contrastive language-audio pretraining that develops audio representations by integrating audio data with natural language descriptions. We fine-tune CLAP to enhance its capture of dance style information. Specifically, we freeze the weights of the audio and text encoders in CLAP and introduce multi-layer perceptrons as adapter layers. These layers are trained using data from the AIST++ dataset. For movement information, we employ Wav2CLIP, initially trained on audio-visual datasets. We apply a similar fine-tuning approach as with CLAP, using the AIST++ dataset to adapt Wav2CLIP for our specific requirements. Regarding rhythm, which is crucial for aligning dance movements with the music beat, we use Librosa, a tool for audio signal analysis, to extract beat information. Given a piece of music $ m $, we pass it through fine-tuned CLAP, fine-tuned Wav2CLIP, and Librosa to extract different features. Specifically, we obtain the music representation $ e_{c} $ from CLAP, $ e_{w} $ from Wav2CLIP, and beat information $ e_{b} $ from Librosa. We then concatenate $ e_{c} $, $ e_{w} $, and $ e_{b} $ to form a comprehensive musical representation $ e $, which is specifically tailored for generating dance videos.

\begin{figure}[h]
  \centering
  \includegraphics[width=\linewidth]{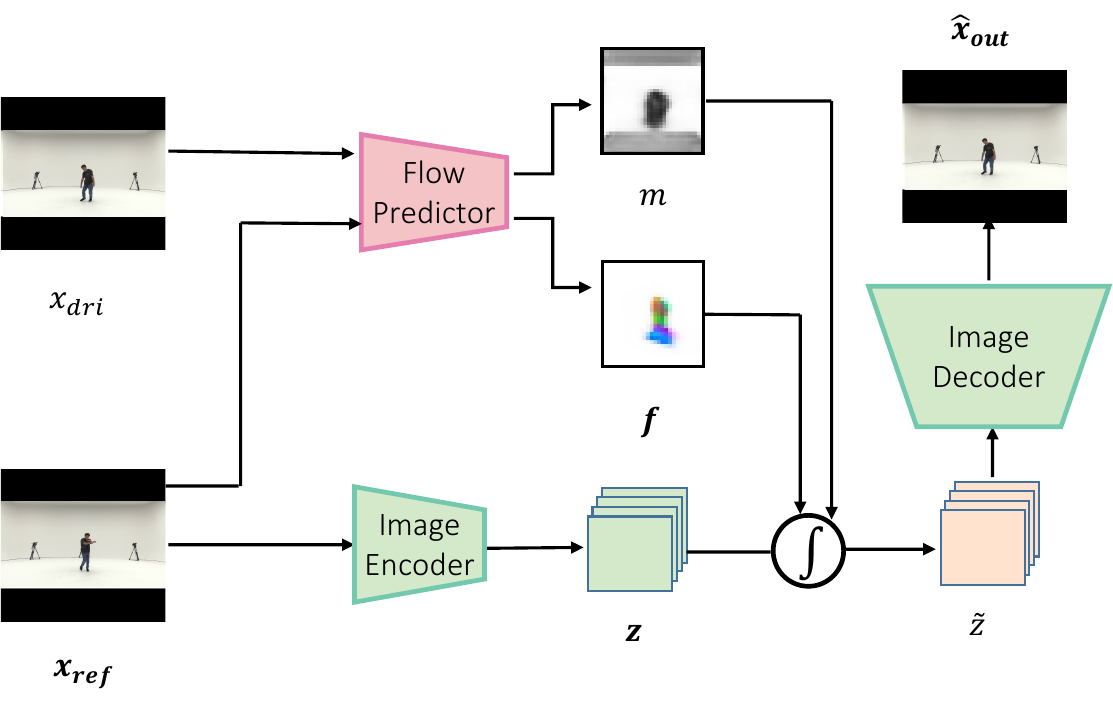}
  \caption{Training of latent flow auto-encoder. The flow predictor learns to estimate the latent flow $ f $ and occlusion map $ m $ between the reference frame $ x_{ref} $ and the driving frame $ x_{dri} $. The image encoder encodes $ x_{ref} $ into a latent representation $ z $, $ f $ and $ m $ are utilized to manipulate $ z $ into $ \tilde{z} $ which is then decoded by an image decoder to generate an output image $ \hat{x}_{out} $. The objective of the training is to minimize the disparity between $ x_{dri} $ and $ \hat{x}_{out} $.}
  \label{fig:stageone}
\end{figure}

\subsection{Latent Flow Estimation}
\label{sec: latent-estimation}
The objective of this phase is the training of an auto-encoder capable of accurately capturing and modeling motion between video frames within a latent space. As illustrated in \cref{fig:stageone}, the architectural components of the model comprise an image encoder, a flow predictor, and an image decoder. During the training process, two frames are randomly selected from a single video to serve as a reference frame $ x_{ref} $ and a driving frame $ x_{dri} $, both sharing dimensions of $ H_{x} \times W_{x} \times 3 $. The image encoder processes $ x_{ref} $, converting it into a compact latent representation, $ z $, with dimensions $ H_{z} \times W_{z} \times C_{z} $. Subsequently, similar to \cite{siarohin2019first, wang2018video}, the flow predictor, receiving both $ x_{ref} $ and $ x_{dri} $ as inputs, computes the backward latent optical flow, $ f $, and an occlusion map, $ m $, to denote the transformation between these frames. The estimated latent flow, $ f $, maintains the spatial dimensions of $ z $ and incorporates two channels to articulate the horizontal and vertical displacement across frames. Using backward flow estimation, the warping of $ z $ using $ f $ is realized through an efficient differentiable bilinear sampling \cite{jaderberg2015spatial}. Concurrently, the occlusion map, sized $ H_{z} \times W_{z} \times 1 $, facilitates the reconstruction of areas within $ z $ that become obscured or revealed due to movement. Values within this map range from 0 to 1, where 1 signifies unoccluded regions, and 0 denotes complete occlusion. The warped latent map, $ \tilde{z} $, is derived through the following equation:
\begin{equation}
  \tilde{z} \ = \ m \ \odot f_{w}(z \ ,\ f),
\end{equation}
where $ f_{w}(\cdot \ , \cdot) $ denotes the back-warping operation and $ \odot $ is the element-wise multiplication. Finally, the image decoder takes $ \tilde{z} $ as input, reconstructing visible portions while simultaneously inpainting occluded regions to produce the output image $ \hat{x}_{out} $. The training objective is to minimize a reconstruction loss that quantifies the difference between $ \hat{x}_{out} $ and $ x_{dri} $. This loss employs the perceptual loss proposed by Johnson et al. \cite{johnson2016perceptual}, which utilizes features extracted by a pre-trained VGG-19 network \cite{simonyan2014very}. Formally, the reconstruction loss is expressed as:
\begin{equation}
  \mathcal{L}_{rec} \ = \ \sum_{i=1}^{N} \left| V_{i}(\hat{x}_{out}) - V_{i}(x_{dri}) \right|,
\end{equation}
where $ V_{i}(\cdot) $ extracts the $ith$ channel features from a specified layer of VGG-19, and $N$ is the total number of feature channels within that layer.

\begin{figure*}[t]
  \centering
  \includegraphics[width=\textwidth]{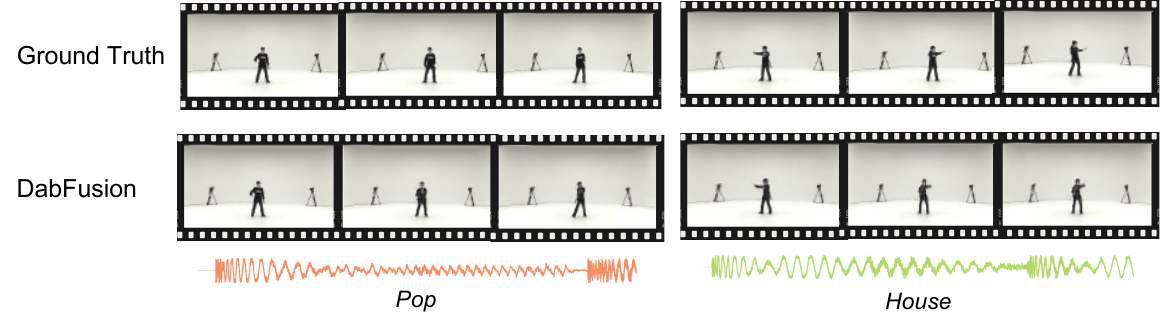}
  \caption{Comparison of video quality between ground-truth videos and those generated by DabFusion. We use the same starting image and music piece to generate videos with our model, and select three frames from the same position.}
  \label{fig:comparison}
\end{figure*}

\subsection{Latent Flow Generation}
\label{sec: diffusion-training}
We use diffusion models to generate latent flow. These generative models reverse a diffusion process, which gradually turns data into Gaussian noise. The reverse process then restores this noise back to the original data distribution. Mathematically, the diffusion process is a Markov chain that adds noise to the data over $ T $ steps. The process starts with the original data $ x_{0} $ and ends with a sample $ x_{T} $ that resembles Gaussian noise. The transition from $ x_{t-1} $ to $ x_{t} $ can be defined by the following equation:
\begin{equation}
    x_{t} = \sqrt{\alpha_{t}}x_{t-1}+\sqrt{1-\alpha_{t}}\epsilon_{t},
\end{equation}
where $ \alpha_{t} $ is a variance schedule that determines the amount of noise added at each step, and $ \epsilon_{t} $ is sampled from a standard Gaussian distribution $ \mathcal{N}(0,\mathcal{I}) $. The reverse diffusion process seeks to reconstruct the original data from noise by learning the conditional distribution $ p(x_{t-1} | x_{t}) $. This is achieved by training a denoising model to estimate the parameters of the Gaussian distribution of $ x_{t-1} $ given $ x_{t} $. The model outputs parameters $ \mu_{\theta}(x_{t},t) $ and $ \sigma_{\theta}(x_{t},t) $, which define the Gaussian distribution from which $ x_{t-1} $ is sampled:
\begin{equation}
    x_{t-1} = \mu_{\theta}(x_{t},t)+\sigma_{\theta}(x_{t},t)\cdot \epsilon,
\end{equation}
where $ \epsilon $ is again sampled from a standard Gaussian distribution. The model is trained to minimize the difference between the generated and actual data distributions, with an objective function often approximating the negative log-likelihood of the data under the model. A practical approach for training diffusion models is employing the Mean Squared Error (MSE) between the predicted and actual noise used in data corruption at each diffusion step. Given the diffusion process, the goal during training is to predict the noise $ \epsilon $ that was added to the original data at each step. The loss function for a single training step can be formulated as:
\begin{equation}
\mathcal{L} = \mathbb{E}_{t, x_0, \epsilon}\left[\|\epsilon - \epsilon_{\theta}(x_t, t)\|^2\right],
\end{equation}
where $ \epsilon_{\theta}(x_t, t) $ is the predicted noise by the model given the noised data $ x_{t} $ at time step $ t $. $ \epsilon $ is the actual noise added to the data $ x_{0} $ to obtain $ x_{t} $.

Our approach utilizes a 3D U-Net \cite{cciccek20163d} as the denoising mode. And the training of the model is in conjunction with the trained image encoder and flow predictor. Given an input video $ x_{0}^{N} = \{ x_{0},x_{1},...,x_{N} \} $ and its corresponding music $ m $, we first use the flow predictor to estimate latent flow sequence $ f_{1}^{N} = \{ f_{1},...,f_{N} \} $ and occlusion map sequence $ m_{1}^{N} = \{ m_{1},...,m_{N} \} $ between starting frame $ x_{0} $ and remaining frames $ \{x_{1},...,x_{N}\} $. The size of $ f_{1}^{N} $ is $ N \times H_{z} \times W_{z} \times 2 $ and the size of $ m_{1}^{N} $ is $ N \times H_{z} \times W_{z} \times 1 $. Subsequently, we concatenate these two sequences along the last dimension to  get $ a_{0} = [f_{1}^{N}, m_{1}^{N} ] $ which has the size $ N \times H_{z} \times W_{z} \times 3 $. Then, we perform diffusion process which gradually adds 3D Gaussian noise to $ a_{0} $ to map it to a a standard Gaussian noise. This is then reversed, using the image encoder to map the initial frame $ x_{0} $ to latent space $ z_{0} $ and incorporating music embeddings $e$ for conditioning the denoising model. The updated loss function, accounting for the conditioning on $ z_{0} $ and $e$, is given by:
\begin{equation}
\mathcal{L} = \mathbb{E}_{t, a_0, \epsilon}\left[\|\epsilon - \epsilon_{\theta}(a_t, t, z_0,e)\|^2\right].
\end{equation}


\section{Experiments and Results}
\subsection{Dataset and Implementation Details}
Our model is trained and evaluated using the AIST++ \cite{li2021ai} dataset, a leading resource in 3D human dance. AIST++ is recognized as the most comprehensive dataset in its domain, encompassing 10 dance genres. Each genre features 6 unique pieces of music, with each piece accompanying multiple videos, totaling 12,670 videos. These videos showcase a range of choreographies from basic to advanced, offering a wide spectrum of movements for detailed analysis. The pivotal advantage of AIST++ for our research is its meticulous synchronization of music and dance movements, which is crucial for music-driven dance generation and analysis. 

DabFusion was trained using two A6000 GPUs. We organize our dataset by music pieces, allocating 10,564 videos and 50 music tracks to the training set, and 2,106 videos with 10 tracks to the test set. Each video is segmented into frames, resized to $ 128\times 128 $ pixels. When fine-tuning CLAP and Wav2CLIP, the adapter layers are composed of two MLP layers, each with a hidden size of 512. These adapter layers are trained over 100 epochs. In the latent flow estimation phase, we adopt the architecture from \cite{johnson2016perceptual} for both our image encoder and decoder. The flow predictor follows \cite{siarohin2019first}. The model is trained over 150 epochs with a batch size of 100, using the Adam optimizer \cite{kingma2014adam}, starting with a learning rate of $ 2\times10^{-4} $, decreased by a factor of 0.1 after epochs 60, 90, and 120. For the denoising model's training, we employ a conditional 3D U-Net architecture from \cite{ho2022video}, featuring four down-sampling and up-sampling 3D convolutional blocks. The time step $ t $ is encoded by the sinusoidal position embedding \cite{vaswani2017attention}. The training spanned 250 epochs, starting with the same initial learning rate as the first stage, and adjusted at epochs 100, 150, and 200. We set $ T $ = 1000 and apply a cosine noise schedule \cite{nichol2021improved} and dynamic thresholding \cite{saharia2022photorealistic} at the 90\% during sampling.

\begin{table}[h]
  \caption{Quantitative results between DabFusion and MM-Diffusion.}
  \label{tab:quantitative comparison}
  \resizebox{\linewidth}{!}{
\begin{tabular}{c|cccc}
    \toprule
    Model&FVD $\downarrow$&LPIPS$\downarrow$&PSNR$\uparrow$&SSIM$\uparrow$\\
    \midrule
    MM-Diffusion & \textbf{180.31}&\textbf{0.025}&\textbf{34.81}&0.951\\
    DabFusion (ours) & 193.98&0.033&26.88&\textbf{0.959}\\
    \bottomrule
\end{tabular}
}
\end{table}

\subsection{Evaluation Metrics}
To assess the quality of generated videos, we use the Fr\'{e}chet Video Distance (FVD) \cite{unterthiner2018towards}. For quantitative assessment of image-level quality, we employ SSIM \cite{wang2004image}, PSNR \cite{hore2010image}, and LPIPS \cite{zhang2018unreasonable}. Additionally, we use the CLIP-Score (CS) \cite{sheffer2023hear} to measure the cosine similarity between CLIP embeddings of the dance video and Wav2CLIP embeddings of the music, evaluating the coherence between the dance video and the music. To evaluate synchronization between the dancer's movements and the music beat, we introduce the 2D motion-music alignment score (2D-MM Align). This metric, inspired by the Beat Alignment Score \cite{li2021ai} for 3D scenes, adapts the concept for 2D environments. The alignment score is calculated as the average distance between each kinematic beat and its closest music beat. 2D-MM Align is defined as:
\begin{equation}
\text{2D-MM Align} = \frac{1}{n} \sum_{i=1}^{n} \exp\left(-\frac{\min_{\forall f_{j}^{y} \in F^y} \, \|f_{i}^{x} - f_{j}^{y} \|^2}{2\sigma^2}\right),
\end{equation}
where $ F^y = \{ f_{j}^{y} \} $ is the music beats, $ F^x = \{ f_{i}^{x} \} $ is the kinematic beats and $ \sigma $ is a normalized parameter. Additionally, we incorporate the Audio-video Alignment Score (AV-Align) \cite{yariv2024diverse} to quantify audio-video synchronization. This metric analyzes optical flow in the video stream and audio peaks in the audio stream and assesses their temporal alignment within a three-frame window.

\begin{figure*}[h]
  \centering
  \includegraphics[width=\textwidth]{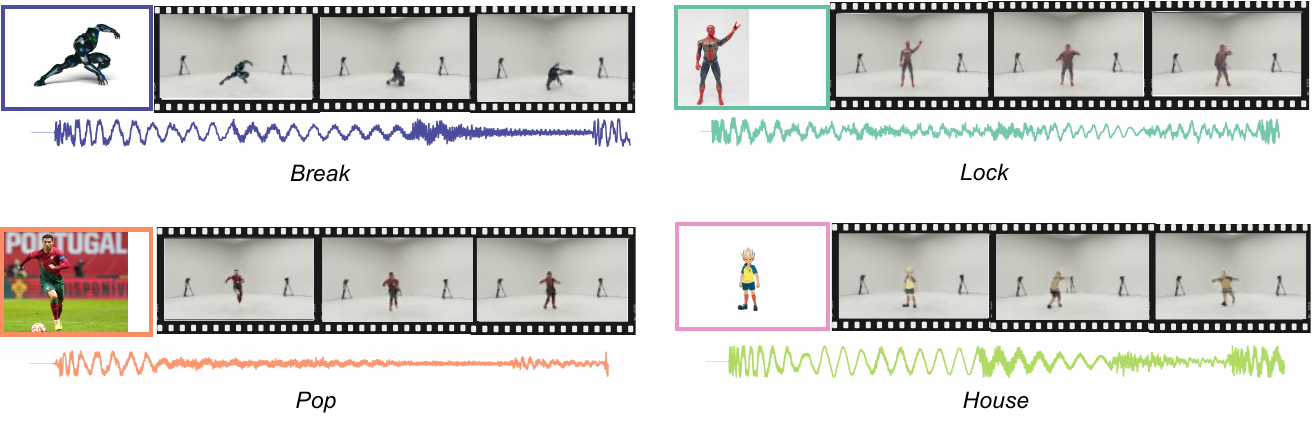}
  \caption{ Exemplar videos generated using our DabFusion, showcasing dance videos with unseen individuals in various poses.}
  \label{fig:unseen}
\end{figure*}

\subsection{Result Analysis}
\noindent \textbf{Video Quality Evaluation}. To evaluate the quality of the videos generated by our DabFusion, we use the music and the first frame from the test set, resulting in 2,106 generated videos. We visually present outputs from our model alongside ground truth in \cref{fig:comparison}. As shown in \cref{fig:comparison}, the quality of the videos generated by our model closely approximates that of real videos. Given the novelty of our task and the lack of directly comparable previous works, we benchmark our DabFusion against MM-Diffusion \cite{ruan2023mm}, the leading model that in unconditional video generation. The AIST++ dataset uniquely utilizes nine cameras ranging from C01 to C09 to record dances from multiple perspectives. DabFusion is trained on the whole dataset and MM-Diffusion is trained on data from only the C01 camera which is in front of the dancer. We randomly generate an equivalent number of videos with MM-Diffusion for comparison. A quantitative comparison between our models and MM-Diffusion is provided in \cref{tab:quantitative comparison}. The results reveal that the quality of videos produced by our models closely matches that of MM-Diffusion. Given that our model can generate videos from various angles, producing videos from specific angles such as C02, which is situated in the upper right of the dance space, proves more challenging than from C01. Despite this, DabFusion demonstrates exceptional performance in the quality of the generated videos.
\begin{table}[h]
  \caption{Quantitative results across different angles. C01 is in front of the dancer, C02 is in the upper right of the dancer, C03 is in the right of the dancer, C09 is in front of the dancer but closer than C01.}
  \label{tab:quantitative comparison different aspects}
  \resizebox{\linewidth}{!}{
 \begin{tabular}{cc|cccc}
    \toprule
    Model&Camera&FVD $\downarrow$&LPIPS$\downarrow$&PSNR$\uparrow$&SSIM$\uparrow$\\
    \midrule
    MM-Diffusion&C01&176.93&\textbf{0.019}&\textbf{35.52}&0.949\\
    \midrule
    \multirow{4}{*}{DabFusion (ours)}&C01& 158.02&0.027&26.33&\textbf{0.964}\\
      &C02& 198.17&0.040&25.23&0.954\\
      &C03& \textbf{157.94}&0.036&26.29&0.960\\
      &C09& 195.17&0.035&25.28&0.956\\
  \bottomrule
\end{tabular}
}
\end{table}
To further assess the impact of camera angles on video quality, we analyze four specific angles. These are summarized in \cref{tab:quantitative comparison different aspects}, illustrating the quality of videos generated from different perspectives. Our models show better performance in terms of FVD and SSIM scores, particularly with the C01 camera position, indicating higher video quality. Our analysis reveals that videos generated from cameras directly facing the dancer, such as C01 and C03, are generally of higher quality compared to those from angled cameras like C02. We also examine the effect of camera distance by comparing videos from C01 and C09, which have the same angle but differ in distance from the dancer. Results indicate that cameras positioned further away tend to produce better outcomes.


\begin{table}[h]
  \centering
  \caption{Alignment results for DabFusion.}
  \label{tab: alignment results}
  \resizebox{\linewidth}{!}{
  \begin{tabular}{c|ccc}
    \toprule
     & CS$\uparrow$ & 2D-MM Align$\uparrow$ & AV Align$\uparrow$ \\
    \midrule
    Ground Truth & 0.153 & 0.223 & 0.152 \\
    DabFusion (ours) & 0.142 & 0.215 & 0.148 \\
    \bottomrule
  \end{tabular}
  }
\end{table}
\begin{table}[h]
  \centering
  \caption{Alignment results based on dance styles.}
  \label{tab:alignment results dance styles }
  \resizebox{\linewidth}{!}{
  \begin{tabular}{cc|ccc}
    \toprule
     Dance Style&Model& CS$\uparrow$ & 2D-MM Align$\uparrow$ & AV Align$\uparrow$ \\
    \midrule
    \multirow{2}{*}{Break}&Ground Truth & 0.138 & 0.231 & 0.226 \\
    &DabFusion (ours) & 0.131 & 0.223 & 0.229 \\
    \midrule
     \multirow{2}{*}{Ballet Jazz}&Ground Truth & 0.063 & 0.198 & 0.101 \\
    &DabFusion (ours) & 0.057 & 0.193 & 0.098 \\
    \midrule
    \multirow{2}{*}{House}&Ground Truth & 0.121 & 0.218 & 0.134 \\
    &DabFusion (ours) & 0.112 & 0.213 & 0.128 \\
    \bottomrule
  \end{tabular}
  }
\end{table}

\noindent\textbf{Alignment Evaluation}. In \cref{tab: alignment results}, we assess the synchronicity between music-video and music-motion pairings. The results indicate that videos generated by our DabFusion closely match the ground-truth videos in terms of alignment metrics. We also evaluate the results according to dance style categories corresponding to the music. These assessments, detailed in \cref{tab:alignment results dance styles }, exhibit significant disparities in alignment scores among various dance genres. Notably, Ballet Jazz yields the lowest alignment scores, attributed to its challenging choreography compared to other dance styles. Consequently, the alignment scores reflect the varying levels of difficulty inherent to each dance genre.

\begin{figure*}[t]
  \centering
  \includegraphics[width=\textwidth]{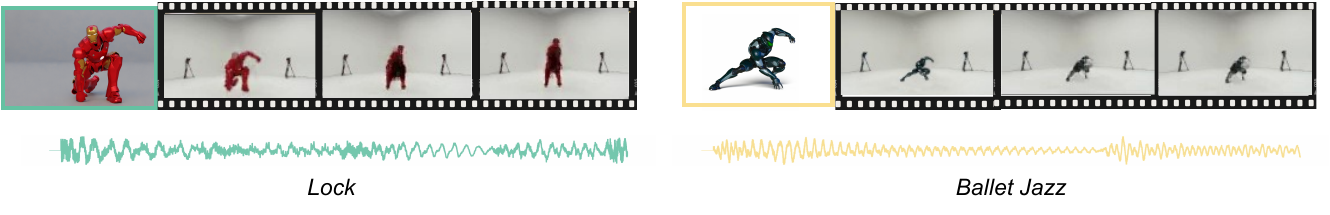}
  \caption{Failure cases for DabFusion in different scenarios. On the left, a person with unclear joint delineations leads DabFusion to produce low-quality video output. On the right, the initial crouching pose, uncommon in Ballet Jazz dance, results in repetitive frames.}
  \label{fig:failure}
\end{figure*}

\subsection{Choreograph Anyone}
DabFusion can choreograph individuals whether they belong to the AIST++ dataset or not, requiring several preprocessing steps. First, we use YOLO \cite{redmon2016you} for object detection and the Segment Anything Model \cite{kirillov2023segment} for segmentation to identify and segment the person from unseen scenarios. We then substitute the individual in the image from the AIST++ dataset with the newly segmented person, creating a combined image. This fused image is set as the starting frame to generate dance videos with musical conditions. We randomly select images featuring diverse individuals in various poses to create dance videos. As illustrated in \cref{fig:unseen}, DabFusion is capable of generating high-quality dance videos featuring previously unseen individuals in a variety of contexts and poses.
\begin{table}[h]
  \centering
  \caption{Ablation study for result comparison between 40-frame video and 80-frame video.}
  \label{tab:ablation_frames}
  \resizebox{\linewidth}{!}{
  \begin{tabular}{c|ccc|c}
    \toprule
    & CS$\uparrow$ & 2D-MM Align$\uparrow$ & AV Align$\uparrow$&FVD $\downarrow$ \\
    \midrule
    40 frames & 0.142 & 0.215 & 0.148 & \textbf{193.98} \\
    80 frames & \textbf{0.153} & \textbf{0.219} & \textbf{0.179} & 194.84 \\
    \bottomrule
  \end{tabular}
  }
\end{table}

\begin{table}[h]
  \centering
  \caption{Ablation study for result comparison between DabFusion with and without beat information.}
  \label{tab:ablation_beat}
  \resizebox{\linewidth}{!}{
  \begin{tabular}{cc|ccc|c}
    \toprule
     Model&Beat& CS$\uparrow$ & 2D-MM Align$\uparrow$ & AV Align$\uparrow$ &FVD $\downarrow$ \\
    \midrule
    \multirow{2}{*}{DabFusion} &$ \times $ &0.138 & 0.202 & 0.141& 194.44 \\
     & $ \checkmark $& \textbf{0.142} & \textbf{0.215} & \textbf{0.148}& \textbf{193.98} \\
    \bottomrule
  \end{tabular}
  }
\end{table}


\subsection{Ablation Study}
\noindent\textbf{Influence of Video Length}. We observe that video length impacts benchmark performance. Our model is trained to generate videos of 40 and 80 frames. Due to training time and computational resource considerations, we designate the 40-frame videos as the baseline. This comparison is detailed in \cref{tab:ablation_frames}. The results indicate comparable video quality between the two lengths; however, the 80-frame videos demonstrate superior performance in alignment evaluations. This improvement is primarily attributed to the extended music duration, providing more information for enhancing alignment accuracy.

\noindent\textbf{Influence of Beat Information}. 
With the rise of large models that generate music representations, many current methods rely solely on these models, often neglecting traditional audio analysis tools. In our experiment, we test the impact of including beat information in our dance video generation. The findings, presented in \cref{tab:ablation_beat}, reveal that while beat information does not greatly change the overall video quality, it improves synchronization between video and audio, and alignment between motion and beat. Specifically, our model that incorporates beat information performs better across all metrics, highlighting the value of including beat information.


\section{Discussion}
Our DabFusion model effectively handles most of the dance generation scenarios; however, it does encounter some limitations in specific situations, as illustrated in \cref{fig:failure}. The first limitation occurs when a dancer's skeleton cannot be clearly distinguished, such as in the left case of \cref{fig:failure}. Additionally, challenges arise when the dancer's starting pose is atypical for the dance style being simulated, as shown in the right case of \cref{fig:failure}. This will lead DabFusion to generate repetitive frames. Moreover, DabFusion is currently limited to creating videos of a fixed length. While chaining sequences together using the final frame of one video as the starting frame for the next can theoretically extend video length, our tests show that this degrades video quality. This deterioration is due to the final frames lacking the fidelity of typical video frames. Future research could improve the quality of longer videos and enable choreography of new dancers without pre-segmentation, making the process simpler and more versatile.



\section{Conclusion}
We present the novel challenge of generating dance videos from music using images of specific individuals, tackled by our Dance Any Beat Diffusion model (DabFusion). It can animate any individual from a single reference image and synchronize their movements to diverse musical styles, demonstrating its versatility and adaptability. Our innovative music encoder is designed to pinpoint critical features such as dance style, movement, and rhythm, which are crucial for dynamic choreography. Employing the AIST++ dataset, we validate DabFusion’s ability to create high-quality dance videos that achieve excellent audio-video synchronization and precise motion-music alignment. This research establishes a solid baseline for future explorations in this intriguing field of dance video generation.

{\small
\bibliographystyle{ieee_fullname}
\bibliography{DabFusion}
}

\end{document}